\documentclass{article}

\PassOptionsToPackage{numbers, sort&compress}{natbib}
\usepackage[preprint]{NeurIPS_2023}

\usepackage[utf8]{inputenc} % allow utf-8 input
\usepackage[T1]{fontenc}    % use 8-bit T1 fonts
\usepackage{hyperref}       % hyperlinks
\usepackage{url}            % simple URL typesetting
\usepackage{booktabs}       % professional-quality tables
\usepackage{amsfonts}       % blackboard math symbols
\usepackage{nicefrac}       % compact symbols for 1/2, etc. 
\usepackage{microtype}      % microtypography
\usepackage{xcolor}         % colors
\usepackage{float}         % colors
\usepackage{algorithm}
\usepackage{algpseudocode}
\usepackage{graphicx}
\usepackage{epstopdf}
\usepackage{amsmath}
\usepackage{multirow}
\usepackage{wrapfig}

\title{RSRM: Reinforcement Symbolic Regression Machine}

\author{Yilong Xu$^1$, Yang Liu$^2$, Hao Sun$^{1,}$\thanks{Corresponding author}  \vspace{3pt} \\
{\small $^1$Gaoling School of Artificial Intelligence, Renmin University of China, Beijing, China;} \\ {\small $^2$School of Engineering Science, University of Chinese Academy of Sciences, Beijing, China;} \vspace{3pt}  \\
{\small Emails: \url{xuyilong88@ruc.edu.cn};~~\url{liuyang22@ucas.ac.cn};~~\url{haosun@ruc.edu.cn}}
}

\begin{document}

\maketitle

\begin{abstract}
In nature, the behaviors of many complex systems can be described by parsimonious math equations. Automatically distilling these equations from limited data is cast as a symbolic regression process which hitherto remains a grand challenge. Keen efforts in recent years have been placed on tackling this issue and demonstrated success in symbolic regression. However, there still exist bottlenecks that current methods struggle to break when the discrete search space tends toward infinity and especially when the underlying math formula is intricate. To this end, we propose a novel Reinforcement Symbolic Regression Machine (RSRM) that masters the capability of uncovering complex math equations from only scarce data. The RSRM model is composed of three key modules: (1) a Monte Carlo tree search (MCTS) agent that explores optimal math expression trees consisting of pre-defined math operators and variables, (2) a Double Q-learning block that helps reduce the feasible search space of MCTS via properly understanding the distribution of reward, and (3) a modulated sub-tree discovery block that heuristically learns and defines new math operators to improve representation ability of math expression trees. Biding of these modules yields the state-of-the-art performance of RSRM in symbolic regression as demonstrated by multiple sets of benchmark examples. The RSRM model shows clear superiority over several representative baseline models.

\end{abstract}

\section{Introduction}

The pursuit of mathematical expressions through data represents a crucial undertaking in contemporary scientific research. The availability of quantitative mathematical expressions to depict natural relationships enhances human comprehension and yields more precise insights. Parsing solutions offer superior interpretability and generalization compared to numerical solutions generated by neural networks. Additionally, simple expressions exhibit computational efficiency advantages over the latter. Consequently, these techniques have found applications in various fields, including physics \cite{aifeynman, liu2021machine, sun2022symbolic} and material sciences \cite{wang2019symbolic}, in recent years.

% However, this task is not without challenges, as nesting mathematical expressions can yield an overwhelming number of possibilities. The exponential growth of mathematical expressions as the number of expression layers increases further complicates the search for interpretable expressions. A straightforward search method often fails to identify the required expressions. Furthermore, due to the large error incurred by even slight changes in mathematical symbols, achieving symbolic learning tasks through optimization becomes inherently difficult, leading computer science theoretical researchers to consider symbolic learning tasks as NP-hard.

The initial process of fitting expressions involves polynomial interpolation to derive an equation, followed by the application of the SINDy method \cite{sparse} utilizing sparse regression to identify appropriate mathematical expressions based on predefined library of candidate terms. These methods effectively reduce the search space from an infinitely large set of possibilities to a limited fixed set of expressions, thereby narrowing down the search process \cite{sun2021physics,chen2021physics,champion2019data}. However, the applicability of this approach is limited, since the compositional structure of many equations cannot be predefined in advance. Therefore, there is a need for more comprehensive methods to search for expressions. %Notably, updated versions of SINDy incorporating parallel computing has recently been introduced \cite{kaheman2020sindy} or improve the method using techniques like bragging \cite{fasel2022ensemble}.

The EQL (Equation Learner) \cite{martius2016extrapolation,sahoo2018learning} model was then introduced as a novel method in symbolic learning. The EQL model addressed the limitations of traditional activation functions used in deep neural networks, by incorporating symbolic operators as activation functions. This modification enabled the neural network to generate more complex functional relationships, allowing for the discovery of intricate math expressions. However, given its compact structure of EQL, optimizing the sparse network to distill parsimonious equation becomes a key challenge. 

Another approach involves generating optimal expression trees \cite{2007Automata}, where internal nodes correspond to operators and each leaf node represents a constant or variable. By recursively computing the expressions of the sub-trees, these expression trees can be transformed into the original math expressions. Initially, genetic programming (GP) \cite{schmidt2009distilling,augusto2000symbolic,improving} were employed to address these problems. Although GP showed promise, its sensitivity to parameter settings leads to instability. Deep learning methods emerged then to tackle the problem. SymbolicGPT \cite{symbolicgpt} utilizes a generative model like GPT to create expression trees, while while AIFeynman \cite{aifeynman} uses neural networks to analyze the relationships and dependencies between variables and search for relevant expressions. An updated version of AIFeynman \cite{aifeynman2} has been developed, offering faster and more precise expression search capabilities. Additionally, reinforcement learning approaches \cite{sun2022symbolic} have been introduced, which utilizes the Monte Carlo tree search method to explore and discover expressions, along with a module-transplant module that generates new expressions based on existing ones. Deep reinforcement learning methods, e.g. DSR \cite{dsr}, utilize recurrent neural networks to learn expression features and generate probabilities. A search algorithm samples these probabilities to generate a batch of expressions, which are subsequently evaluated for performance. The network is then trained using these evaluations. Combining DSR and GP leads to a new model called NGGP \cite{nggp} achieved better performance. Additionally, pre-trained generative models \cite{dgsr} and end-to-end transformer modules \cite{kamienny2022end,transformerbased} have been also employed to achieve improved expression search results.

Nevertheless, the existing methods still struggle with generating lengthy and complex equations, and are faced with issues related to overfitting. To overcome these challenges, we propose a model named Reinforcement Symbolic Regression Machine (RSRM) that masters the capability of uncovering complex math equations from only scarce data. The search strategy in our model is based on the synergy between double Q-learning \cite{hasselt2010double} and Monte Carlo tree search \cite{Coulom2007Efficient}. By employing two reinforcement methods without deep learning, we aim to mitigate overfitting concerns. To capture the distribution of equations, we utilize double Q-learning, while Monte Carlo Tree Search (MCTS) aids in generating new expressions. Additionally, to address the challenge of lengthy and hard equations, we introduce an interpolation method to identify whether the equation exhibits symmetry prior to each search, followed by a modulated sub-tree discovery block. If symmetry is present, we pre-process the equation accordingly to simplify the subsequent search process. This approach effectively reduces the difficulty associated with specific equations. The sub-tree form-discovery module is proposed to examine whether the few expressions that perform well adhere to a specific form. For instance, if both $e^x-x$ and $e^x+x$ yield favorable results, the expression can be confirmed as $e^x+f(x)$, thereby allowing us to focus on finding $f(x)$. This divide-and-conquer algorithm enables a step-by-step search for equations, facilitating the generation of long expressions.

% The main contributions of our model include exploring the properties of individual variables within the equation, such as symmetry. Additionally, our approach involves determining the form of the equation initially, followed by further generation steps. To address the challenges posed by long equations, we employ the form-discover algorithm, which identifies common patterns among expressions that exhibit satisfactory performance. This allows us to narrow down the search space by focusing on specific forms, thereby facilitating the generation of lengthy expressions.

The main contributions of this paper are presented as follows. Our proposed RSRM model offers a novel solution to the search for mathematical expressions. By incorporating double Q-learning and Monte Carlo tree search, we effectively address issues of overfitting and generate new expressions. Through the utilization of modulated sub-tree discovery block, we handle equations with symmetry, reducing their complexity. Furthermore, the block also assists in dealing with long equations by identifying common patterns. By following this comprehensive approach, we enhance the efficiency and effectiveness of the expression search process. As a result of these advancements, the RSRM model demonstrates clear superiority over several representative baseline models. Its state-of-the-art performance surpasses that of the baseline models in terms of accuracy and generalization ability.

\section{Background}

\textbf{Genetic Programming:} Genetic programming \cite{gplearn,genetic,schmidt2009distilling} is employed to iteratively improve expression trees in order to approximate the optimal expression tree. The mutation step in GP enables random mutations in the expression tree, while genetic recombination allows for the exchange of sub-trees between expression trees, leading to the creation of new expression trees based on the knowledge acquired from previous generations. This ``genetic evolution'' process progressively yields highly favorable outcomes after a few generations.

\textbf{Double Q-Learning:} Double Q-learning \cite{hasselt2010double} is a variant of Q-learning that employs two separate Q-value functions. During training, one structure's results are used to train the other structure with a random 50\% probability. During decision-making, both models collaborate to make decisions. This approach mitigates overestimation issues by ensuring that Q-value updates are not solely based on values used for action selection. As a result, more accurate Q-value estimation is achieved, leading to improved agent performance.

\textbf{Monte Carlo Tree Search:} MCTS \cite{Coulom2007Efficient} is a decision-making search algorithm that constructs a search tree representing possible game states and associated values. It employs stochastic simulations to explore the tree and determine the value of each node. This algorithm gained prominence via its adoption by the AlphaZero team \cite{silver2017mastering}. MCTS consists of four steps in each iteration: (1) selection, (2) expansion, (3) simulation, and (4) backpropagation. During selection, the best child node is chosen based on certain criteria. If an expandable node lacks children, it is extended by adding available children. The simulation step involves simulating the current state before selecting the next node, often using the Upper Confidence Bound for Trees (UCT) algorithm to calculate the selection probabilities. Finally, in the backpropagation step, the reward function evaluates child nodes, and their values are used to update the values of parent nodes in the tree.
\begin{equation}
    UCT(n) = \bar{Q}(v')+c\sqrt{\frac{\ln(N(v))}{N(v')}}.
    \label{eq:UCT}
\end{equation}
Here, $\bar{Q}(v')$ means the average reward of child node, $N(v)$ and $N(v')$ means the visit time of node and its child. The first part makes the nodes with high expectations visit more often, and the second part ensures that the nodes with fewer visits have a higher probability of being selected.

\section{Method}

\begin{wrapfigure}[18]{r}{0.65\textwidth}
\vspace{-47pt}
\begin{center}
\includegraphics[width=0.99\linewidth]{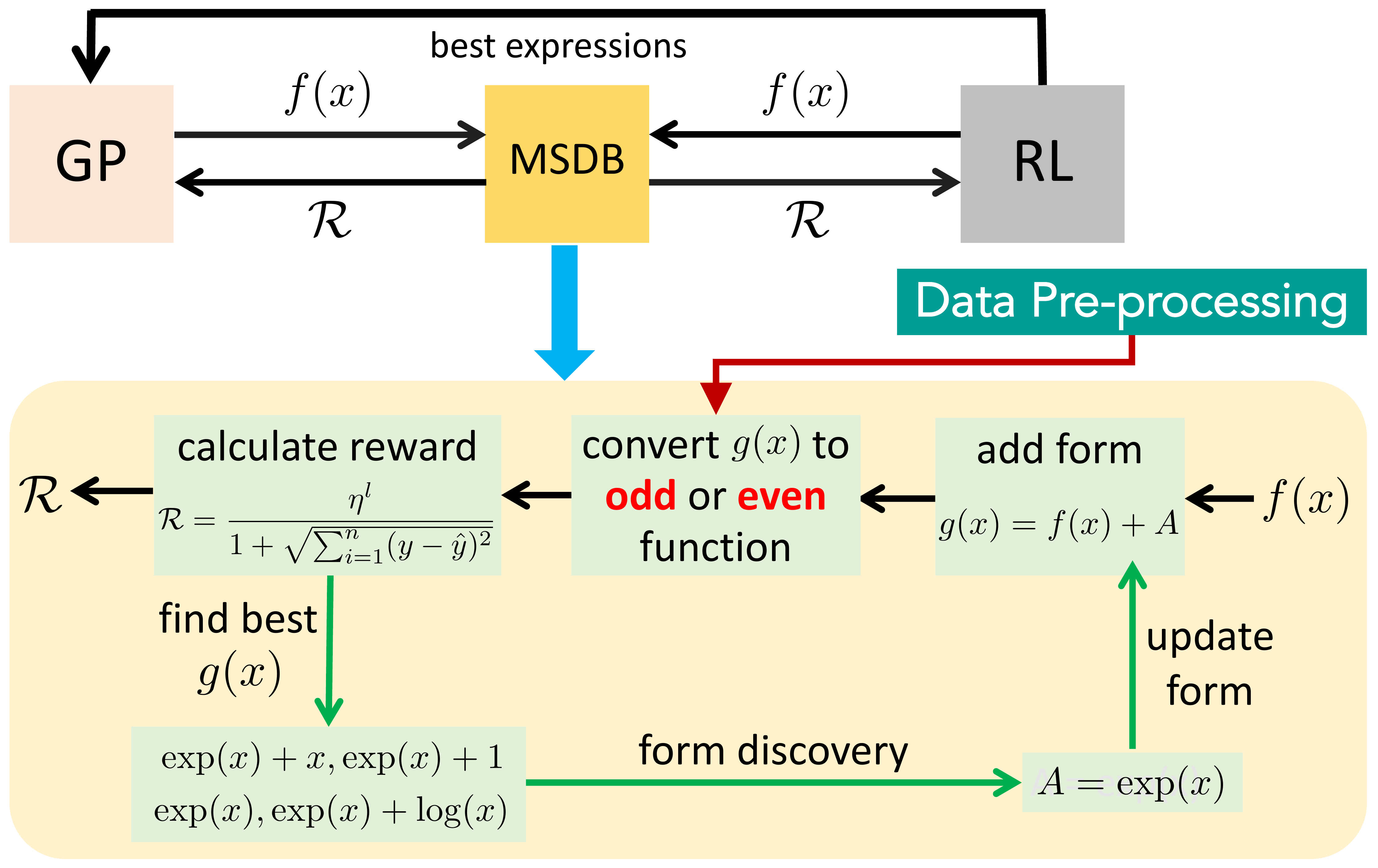}
\vspace{-6pt}
\caption{The proposed RSRM utilizes a schematic flow comprising several steps. First, the function is put into the data pre-processing module to determine the parity, followed by a RL-guided search in the Modulated Sub-tree Discovery Block (MSDB) to manipulate expressions and evaluate their form. The discovered equations are then refined using GP. MSDB is continuously updated based on the identified equations. }
\label{fig:overall}
\end{center}
\vspace{-16pt}
\end{wrapfigure}
\normalsize

The proposed RSRM model consists of a three-step symbolic learning process: reinforcement learning-based expression search, GP tuning, and modulated expression form discovery. Through these steps, our model effectively learns and represents the relationship present in the data, facilitating accurate and interpretable modeling. The schematic representation of RSRM is depicted in Figure \ref{fig:overall}. Full settings of our model are in Appendix \ref{sec:model}.

%We now provide an overview of the entire expression search process. Firstly, expressions are initially generated using the reinforcement learning method. The generated expressions that demonstrate better performance are then subjected to fine-tuning using the genetic algorithm \cite{deap}. After several rounds of generation and refinement, the confirmed part of the expression is fixed, and a new reinforcement learning agent (with non-inherited parameters) is employed to carry out the subsequent generation step, encompassing expression search, tuning, and parameter updates. Each time a newly generated expression is introduced, it is incorporated into the slot to generate a new expression. Following a few rounds of exploration using the new agent, the exploration of the old agent is resumed to identify any potential improved slots. This cycle continues until the best expression is discovered. Full settings of our model are in Appendix \ref{sec:model}.

\subsection{Expression Tree}
In the context of symbolic learning tasks, the underlying objective is to transform the given task into the generation of an optimal expression tree \cite{2007Automata}, which represents a mathematical expression. The expression tree consists of \textit{internal nodes} that correspond to operators (e.g., $+, -, \times, \div, \log, \exp, \sin, \cos$) and \textit{leaf nodes} that correspond to constants (e.g., $1, 2$) or variables (e.g., $x$). By recursively computing the expressions of the sub-trees, the expression tree can be transformed into the original mathematical expressions. The process of generating an expression tree follows a recursive method where operators are added until no more can be added. This approach simplifies the task of creating expressions as it focuses on constructing the expression tree, which can be easily generated using recursive techniques. By adopting the concept of expression trees, symbolic learning tasks can be effectively addressed by generating and manipulating these trees, enabling the representation and computation of complex mathematical expressions.

%This approach allows for the symbolic representation of the task, where the structure of the expression tree captures the relationships and operations involved in the problem. By generating and manipulating expression trees, symbolic learning algorithms aim to discover and represent the underlying mathematical patterns or equations that govern the task at hand. The transformation from expression trees to math expressions enables a clear interpretation of the learned knowledge and facilitates further analysis and utilization of the discovered mathematical relationships.

In contrast to previous methods, we employ a hierarchical traversal strategy for generating expression trees. This is motivated by the Monte Carlo tree search algorithm, where conducting more searches on vertices that are filled earlier is deemed more beneficial. In the context of constructing expression trees, this implies that higher-level nodes in the tree carry greater significance. Consequently, we adopt a layer-by-layer search method, ensuring that higher-level regions of the tree receive a higher number of search iterations. By prioritizing the traversal of higher-level nodes, we place emphasis on exploring and refining the more influential components of the expression tree. This hierarchical approach allows us to effectively capture and exploit the hierarchical structure of the problem, potentially leading to improved performance and more meaningful results.

\begin{figure}[t!]
    \centering
    \includegraphics[scale=0.15]{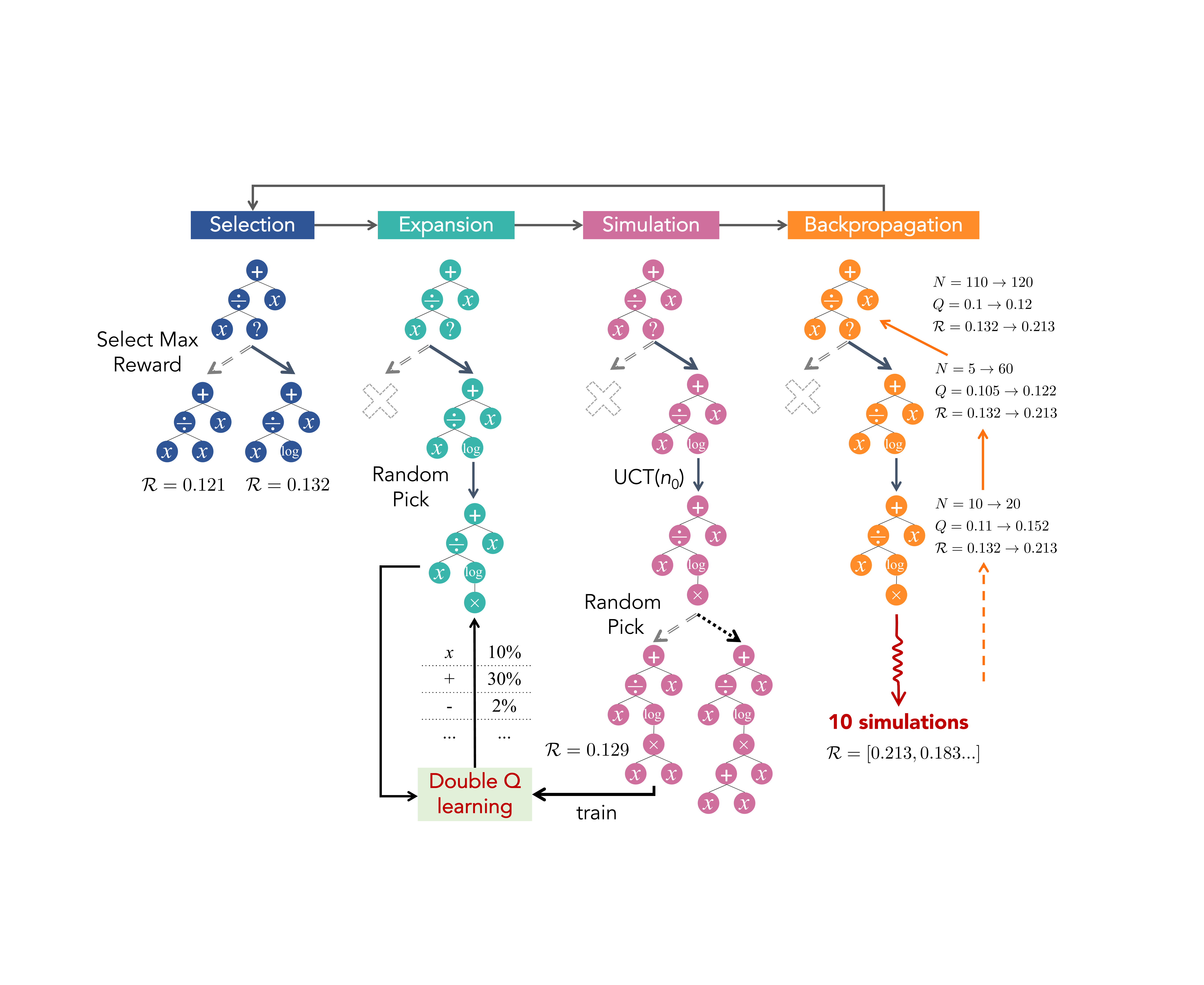}
    \caption{Schematic of the proposed RL search. MCTS selects functions based on the maximum reward, expands them using the results of Q-learning, simulates node selection through the UCT function, randomly fills the current tree, and provides rewards to double Q-learning. Once the generation is complete, the rewards are back-propagated to the parent node. }
    \label{fig:search}
\end{figure}

\subsection{Reinforcement Learning Guided Search}
The search step relies on the double Q-learning and MCTS algorithms, which are shown in Figure \ref{fig:search}. The specific algorithm is shown in Algorithm \ref{alg:search}.

\textbf{Reward function}: The reward function used in our approach is based on the root mean square error (RMSE) and is designed to evaluate the fit of the generated equations to the measured data. It promotes concise and accurate expressions by assigning higher rewards to shorter and more precise functions. The reward function is computed as follows:
\begin{equation}
\mathcal{R} = \frac{\eta^l}{1+\sqrt{\sum_{i=1}^n (y_i-\hat{y_i})^2}},
\label{eq:reward}
\end{equation}
where $\eta$ denotes a discount factor that promotes concise results, $l$ the number of nodes in the expression tree, and $y_i$ and $\hat{y_i}$ the true value and the predicted value generated by the MSDB with the output of Reinforcement Learning Search of the $i$th data point, respectively. By using this reward function, our approach encourages the discovery of equations that minimize the RMSE and favors shorter and more concise expressions, leading to higher reward values for functions that provide better fits to the data.

\begin{algorithm}[t!]
%\small
    \caption{Expression generation by RSRM}\label{alg:search}
    \begin{algorithmic}
        \renewcommand{\algorithmicrequire}{\textbf{Input:}}
        \Require dataset $\mathcal{S}_{data}$, current expression form $\mathcal{F}$
        \renewcommand{\algorithmicrequire}{\textbf{Parameters:}}
        \Require discount rate $\eta$, training rounds $t_r$, UCT const $c$, minimum selected times $n_0$
        \renewcommand{\algorithmicrequire}{\textbf{Outputs:}}
        \Require best expression 
        \State Initiate $S$ as top of MTCS
        \State \textbf{Selection}:
        \State $a \gets$ children of $S$ with maxium $\mathcal{R}$ {\color{gray}\Comment{\small Greedy selection}}
        \State $S$ take action $a$
        \State \textbf{Simulation}:
            \State $S'\gets S$
            \Repeat
                \If {children of $S$ is empty} Expand $S'$ \EndIf
                \If {$\exists x \in $ children of $S' \rightarrow (N(x) < n_0)$} $a' \gets x$ {\color{gray}\Comment{\small Select child with visit times < $n_0$}}
                \Else \quad $a' \gets$ randomly choose child of $S'$ by $UCT$
                {\color{gray}\Comment{\small Select through UCT function}}
                \EndIf
                \State$S'$ take action $a'$, $S'' \gets S'$, Fill up randomly $S''$
                \State double Q-learning $\gets S', a', \mathcal{R}$ of $S''$ 
                {\color{gray}\Comment{\small train double Q-learning by simulated reward}}
            \Until $S'$ is full 
        \State \textbf{Expansion}
        \State children of $S\rightarrow$ double Q-learning $\rightarrow p_{children}$ 
        {\color{gray}\Comment{\small estimate initial possibility of each child}}
        \State \textbf{Back-propagate} $\mathcal{R}$ of $S'$ based on $\mathcal{F}$
    \end{algorithmic}
\end{algorithm}

\textbf{Greedy selection}: Our method employs greedy selection, similar to SPL \cite{sun2022symbolic}. Instead of selecting the token with the highest UCT (Eq. \ref{eq:UCT}) score, we choose the token that currently yields the best reward (Eq.  \ref{eq:reward}). This ensures the selection of tokens leading to expressions resembling the current best one, potentially resulting in improved expressions.

\textbf{Simulated reward}: At each token generation, the entire expression tree is randomly completed based on the current tree. The reward is then computed using the reward function and fed back to double Q-learning for training. This approach avoids excessive rounds of learning at the top node and filters out irrelevant nodes initially.

\textbf{Parameter optimization}: Parameter placeholders are used in expressions, treating each parameter as an unknown variable. Real variables are substituted into the original equation, and the parameters are optimized to minimize the error between predicted value and real value. The BFGS \cite{bgfs} algorithm, available in the scipy \cite{scipy} module in Python, is used for optimization. Gaussian random numbers with unit mean and variance serve as better starting points for optimization.

\subsection{Modulated Sub-tree Discovery}
%\textbf{Expression constraint}: In this paper, we use the prior constraint in DSR\cite{dsr} to reduce the search space. (1) The expressions are restricted to pre-specified minimum and maximum lengths. If the current length is less than the minimum length, variables $x, y. . . $ and parameters will not be generated, and if the current length plus the number of nodes to be generated equals the maximum length, then only these two nodes will be provided. (2) The successor node of a unary operator should not be the inverse of that operator. (4) The successor node of a trigonometric function node should not be a trigonometric function. (5) The maximum number of parameters specified. 

Our approach incorporates three specific search forms to enhance the exploration and analysis of equations, where $\mathcal{A}$ represents a fixed form and $f(x)$ represents a learnable part. The three forms are explained as follows:
\begin{itemize}
    \item $\mathcal{A} \pm f(x)$: This search form focuses on identifying expressions of the form like $e^x - x$ and $e^x + x$. By recognizing this pattern, we can effectively explore and analyze equations that follow the structure of $e^x \pm f(x)$.
    
    \item $\mathcal{A}\times f(x)$: In this search form, we target expressions such as $1.57 e^x$ and $1.56e^x + x$, aiming to detect equations of the form $e^x \times f(x)$.
    
    \item $\mathcal{A}^{f(x)}$: The search form $\mathcal{A}^{f(x)}$ is designed to recognize equations like $(e^x)^{2.5}$ and $(e^x)^e$, indicating the presence of expressions in the form $(e^x)^{f(x)}$.

\end{itemize}
The complete from-discovery algorithm, which outlines the procedure for selecting and generating the search form among the three options, is provided in Algorithm \ref{alg:form}. 
\begin{algorithm}[t!]
%\small
\caption{Search for the form of the expression through the generated expressions}\label{alg:form}
    \begin{algorithmic}
        \renewcommand{\algorithmicrequire}{\textbf{Input:}}
        \Require symbol set $\mathcal{S}_{sym}$, best expression set $\mathcal{S}_{best}$
        \renewcommand{\algorithmicrequire}{\textbf{Parameters:}}
        \Require discount rate $\eta$, selection ratio $k_s$, expression percentage ratio $k_p$, maximum select number $N$
        \renewcommand{\algorithmicrequire}{\textbf{Output:}}
        \Require the form of the expression $\mathcal{F}$
        \State $l \gets$ length of $(\mathcal{S}_{best})$
        \State Sort $\mathcal{S}_{best}$ by $\mathcal{R}$ decent 
             \For{$i$ in $1, 2 ... l$ }
                \If{$i\le N $ and $\mathcal{R}(\mathcal{S}_{best}[i])\ge k\times \mathcal{R}_{max}$}
                {\color{gray}\Comment{\small If number of $\mathcal{G}$ exceeds or $\mathcal{R}$ is low, break out}}
                
                     \State $\mathcal{D} = \mathcal{D}+\text{Split}(\mathcal{S}_{best}[i])$
                {\color{gray}\Comment{\small Occurrences of function in Split-by-addition($\mathcal{S}_{best}[i]$) $+1$}}
                \EndIf
            \EndFor
        \State $\mathcal{G}_0 \gets \mathcal{D}$ with maximum number of occurrences.
        \If{$\exists \mathcal{C} \notin Z \rightarrow \mathcal{G}_0 = \mathcal{A}^{\mathcal{C}} $}  $\mathcal{F} = \mathcal{A}^ {f(x)}${\color{gray}\Comment{\small The form is $\mathcal{A}^{f(x)}$}}
        \ElsIf{$\exists \mathcal{C} \notin Z \rightarrow \mathcal{G}_0 = \mathcal{A}\times \mathcal{C} $} $\mathcal{F} = \mathcal{A} \times f(x)${\color{gray}\Comment{\small The form is $\mathcal{A}\times f(x)$}}
        \Else
             \State$\mathcal{F} = f(x)${\color{gray}\Comment{\small The form is $\mathcal{A}\pm F(x)$}}
            \For{$\mathcal{G}$ in $\mathcal{D}$}
                \If{Occurrences of $\mathcal{G} \ge l \times k_p$}
                    $\mathcal{F} = \mathcal{F}+\mathcal{G}${\color{gray}\Comment{\small Add $\mathcal{G}$ to $\mathcal{A}$}}
                \EndIf
            \EndFor
        \EndIf
    \end{algorithmic}
\end{algorithm}

\textbf{Splitting by Addition:} In this step, we perform the following process: Convert the formula, which is represented as a token set, into a string expression using a library like sympy \cite{meurer2017sympy}. Expand the expression into a sum of simpler expressions. Split the expanded expression into multiple simple expressions using sum or difference notation. Store these simple expressions in a dictionary for subsequent analysis and computation, while also keeping track of the count for each simple expression. This process allows us to break down the formula into its constituent parts, making it easier to work with and analyze individual expressions separately.

These search forms enable our approach to classify equations into different structural patterns, facilitating more focused and efficient exploration. By incorporating RL or GP techniques with the MSDB, we can dynamically modify the $f(x)$ part in each search form and generate equations $g(x)$ that conform to the identified patterns.

We then introduce a data pre-processing module determine the potential parity of the underlying equation. The cubic spline \cite{catmull1974class} is applied for equation fitting, generating a function. Subsequently, this function is utilized to compute the relationship between $y(-x)$ and $y(x)$, enabling determination of whether $y(x)$ is an odd, even, or neither function, where $y(x)$ means the relation of $x$ and $y$ in given data.

When the error between $y(-x)$ and $y(x)$ remains below $E_{sym}$, the function is considered even with respect to $x$. Negative values of the independent variables are transformed to their absolute values, while retaining the dependent variable values. Further exploration is conducted using the form of $\hat y = (g(x) + g(-x))/2$ to make  it to discover specific forms.

Similarly, if the error between $y(-x)$ and $y(x)$ is within the limit $E_{sym}$, the function is classified as odd relative to $x$. Negative values of the independent variables are converted to absolute values, and the dependent variable values are inverted. The search continues employing the form of $\hat y = (g(x) - g(-x))/2$ to make it odd in discover specific forms.

After that we evaluate their performance based on the reward function (see Eq. \ref{eq:reward} ), then the reward values are fed back to the reinforcement learning (RL) or genetic programming (GP) algorithms. This comprehensive approach enhances our ability to explore and analyze a wide range of underlying mathematical expressions in spite of intricate structure.

\section{Results}

\subsection{Experiment on Basic Benchmarks}

In our comparison, we include the following baseline methods in symbolic learning:
\begin{itemize}
\item  \textbf{SPL}\cite{sun2022symbolic}: It utilizes prior knowledge and the MCTS algorithm for symbolic learning.

\item  \textbf{NGGP}\cite{nggp}: An upgraded version of DSR \cite{dsr}, it employs risk-seeking strategy gradient training in deep reinforcement learning, along with GP for optimization.

\item  \textbf{gplearn}\cite{gplearn}: Considered the most stable implementation of symbolic learning using GP.
\end{itemize}

The full settings of baselines are given in Appendix \ref{sec:baseline}. To evaluate the efficiency of our algorithm, we utilize four different benchmarks: 
\begin{itemize}
\item  \textbf{Nguyen} \cite{uy2011semantically}: A standard benchmark for symbolic learning with one or two independent variables and equations randomly sampled over a range of 20-100 data points.

\item  \textbf{Nguyen$^c$} \cite{mcdermott2012genetic}: A parametric version of the Nguyen benchmark, allowing the use of parametric optimization to test equations with parameters.

\item  \textbf{R} \cite{mundhenk2021symbolic}: Consists of three built-in rational equations with numerous polynomials as divisors and divisees, increasing the learning difficulty.

\item  \textbf{LiverMore} \cite{mundhenk2021symbolic}: Contains challenging equations rarely encountered in symbolic learning, including high exponentials, trigonometric functions, and complex polynomials.

\end{itemize}
For evaluation, we employ the recovery rate as a metric, which measures the number of times the correct expression is recovered across multiple independent repetitions of a test. This metric ensures that the model's output exactly matches the target expression.

\begin{table*}[b!]
\centering
  \caption{Recover rate (\%) of several difficult equations in symbolic regression: trigonometric functions and sum of multiple power functions with parameter $1/2$ in Nguyen; power functions and trigonometric functions in Nguyen$^c$, trigonometric functions and hyperbolic function and functions with weird power in LiverMore; rational functions in R. }
  \vspace{6pt}
  \footnotesize
  \begin{tabular}{ccccccc}
    \toprule
    BenchMark&Equation&Ours&SPL&NGGP&DSR&GP\\
    \midrule
    Nguyen-5&$\mathrm{sin}(x_1^2)\mathrm{cos}(x_1)-1$&\textbf{100}&95&80&72&12\\
    Nguyen-12&$x_1^4-x_1^3-0.5x_2^2+x_2$&\textbf{100}&28&21&0&0\\
    \midrule
    Nguyen-2$^\mathrm{c}$&$0.48x_1^4+3.39x_1^3+2.12x_1^2+1.78x_1$&\textbf{100}&94&98&90&0\\
    Nguyen-9$^\mathrm{c}$&$\mathrm{sin}(1.5x_1)+\mathrm{sin}(0.5x_2^2)$&\textbf{100}&96&90&65&0 \\
    \midrule
    LiverMore-3&$\mathrm{sin}(x_1^3)\mathrm{cos}(x_1^2)-1$&\textbf{55}&15&2&0&0\\
    LiverMore-7&$\mathrm{sinh}(x_1)$&\textbf{100}&18&24&3&0\\
    LiverMore-16&$x_1^{2/5}$&\textbf{100}&40&26&10&5\\
    LiverMore-18&$\mathrm{sin}(x_1^2)\mathrm{cos}(x_1)-5$&\textbf{100}&80&33&0&0\\
    \midrule
    R-1$^0$ &$(x_1+1)^3/(x_1^2-x_1+1)$&\textbf{49}&0&2&0&0\\
    R-2$^0$&$(x_1^5-3x_1^3+1)/(x_1^2+1)$&\textbf{89}&0&0&0&0\\
    R-3$^0$&$(x_1^5+x_1^6)/(x_1^4+x_1^3+x_1^2+x_1+1)$&\textbf{91}&0&4&0&0\\
    \bottomrule
\end{tabular}\\
\label{table:hard-eq}
\end{table*}

\begin{wrapfigure}[10]{r}{0.7\textwidth}
\vspace{-24pt}
\begin{center}
\includegraphics[width=0.99\linewidth]{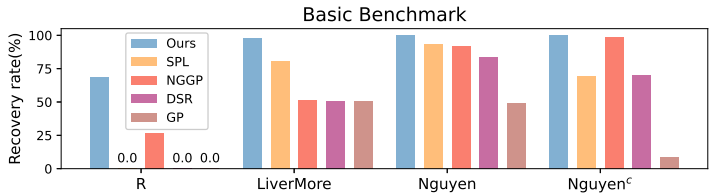}
\vspace{-10pt}
\caption{Recover rate (\%) of basic benchmarks. Detailed results of each each benchmark can be found in Appendix \ref{sec:basic dataset}.}
\label{fig:basic benchmarks}
\end{center}
\vspace{-16pt}
\end{wrapfigure}
\normalsize

We performed a comparison of difficult expressions on the four benchmarks listed in Table \ref{table:hard-eq}. The results demonstrate that our model performs well on these complex expressions. Our model has the ability to break down complex expressions into stepwise simple expression searches using formal search techniques. For example, the equation $x_1^4-x_1^3-0.5x_2^2+x_2$ can be decomposed into the sum of $-0.5x_2^2$ and $x_1^4-x_1^3+x_2$. By distilling the equation into the form $x_1^4-x_1^3+x_2+f(x_1, x_2)$ by the results of RL and genetic algorithms, the subsequent search becomes simpler. As a result, our model achieves a 100\% recovery rate for these expressions.

Furthermore, we compared the mean recovery rates of all equations on each benchmark (see Figure \ref{fig:basic benchmarks}). Our method outperforms other approaches, achieving the highest recovery rates for all expressions.

\subsection{Free-falling Balls Dataset}

We conducted an experimental evaluation on the free-falling balls dataset to assess the parametric learning capability of our model. The dataset consisted of experimental data of balls dropped from a bridge, as described in a previous study \cite{de2020discovery}. The dataset comprised 20-30 observations of ball throw heights within the first 2 seconds, aiming to learn the equation governing ball drop and predict the height between 2 and 3 seconds. Since an exact solution for this dataset is not available, we employed the mean squared error (MSE) as our evaluation metric.

\begin{wraptable}[17]{r}{0.72\textwidth}
\vspace{-18pt}
\caption{MSE of free-falling balls dataset. More detailed data on the equations generated by different models can be found in Appendix \ref{sec:phy}.} \label{table:balls}
\centering
\vspace{6pt}
{\footnotesize
  \begin{tabular}{ccccccc}
    \toprule
    BenchMark&Ours&Ours$^*$&SPL&M-A&M-B&M-C\\
    \midrule
    baseball&\textbf{0.053} &0.068 &0.300 &2.798 &94.589 &3.507\\
    blue basketball&\textbf{0.008} &0.027 &0.457 &0.513 &69.209 &2.227\\
    bowling ball&0.014 &0.034 &\textbf{0.003} &0.33 &87.02 &3.167\\
    golf ball&\textbf{0.006} &0.041 &0.009 &0.214 &86.093 &1.684 \\
    green basketball&0.094 &\textbf{0.045} &0.088 &0.1 &85.435 &1.604\\
    tennis ball&0.284 &\textbf{0.068} &0.091 &0.246 &72.278 &0.161\\
    volleyball&0.033 &\textbf{0.025} &0.111 &0.574 &80.965 &0.76\\
    whiffle ball 1&\textbf{0.038} &0.660 &1.58 &1.619 &65.426 &0.21 \\
    whiffle ball 2&\textbf{0.041} &0.068 &0.099&0.628 &58.533 &0.966 \\
    yellow whiffle ball&1.277 &1.080 &\textbf{0.428}&17.341 &44.984 &2.57 \\
    orange whiffle ball&\textbf{0.031} &0.368 &0.745&0.379 &36.765 &3.257 \\
    \midrule
    \textbf{Average }&\textbf{0.173} &0.242 &0.356 &2.24&71.02&1.828\\
    \bottomrule
\end{tabular}\\}
\normalsize
\end{wraptable} 
We consider two sets of RSRM models, the standard one and the one (named RSRM*) that follows the expression form $c_4x^3+c_3x^2+x_2x+c_1+f(x)$. We compared these models with the baseline method SPL \cite{sun2022symbolic}, since other models like NGGP \cite{nggp}, GP \cite{gplearn}, DGSR \cite{dgsr}, and AIFeynman \cite{aifeynman2} tend to have large generalization errors due to the limited data points (20-30 per training set) in the falling balls benchmark given the fact that the exact solution is unknown.

Three physics models derived from mathematical principles were selected as baseline models for this experiment, and the unknown constant coefficient values were estimated using Powell's conjugate direction method. The equations of the baseline models are presented as follows. \textbf{M-A}: $h(t) = c_1t^3+c_2t^2+c_3t+c_4$, \textbf{M-B}: $h(t) = c_1\exp(c_2t)+c_3t+c_4$, and \textbf{M-C}: $h(t) = c_1\log(\cosh(c_2t))+c_3$.

The results (see Table \ref{table:balls}) show that in most cases, the RSRM model performs better than SPL. The RSRM model can successfully find the equation of motion for uniformly accelerated linear motion ($c_1x^2+c_2x+c_3+f(x)$) and search for additional terms to minimize the training error. This leads to improved results compared to SPL. However, there are cases where RSRM makes mistakes, such as obtaining expressions in the form of $c_1\cos(x)^2+c_2+f(x)$ when searching for the yellow whiffle ball. This increases the generalization error and reduces the overall effectiveness compared to SPL. Overall, RSRM outperforms SPL in physics equation discovery, demonstrating its effectiveness in solving parametric learning tasks on the free-falling balls dataset.

\subsection{Generalization Performance Test}

To compare the generalization ability of our model with other methods, we conducted an experiment on generalization performance. The dataset was generated using the cumulative distribution function (CDF) defined in as follows, with varying means ($\mu$) and variances ($\sigma$). The dataset consisted of 201 points spanning the range from $-$100 to 100.
\begin{equation}
    F(x,\mu,\sigma) = \int_{-\infty}^x \frac{1}{\sqrt{2\pi \sigma}}e^{\frac{-(t-\mu)^2}{2}}dt.
    \label{eq:cdf}
\end{equation}

Considering the limited capability of the SPL model \cite{sun2022symbolic} to learn very complex equations, we selected NGGP \cite{nggp} as the appropriate baseline for evaluating generalization ability. In addition to our model, we included several baselines for comparison: cubic splines \cite{dyn2009four}, NGGP \cite{nggp}, and Multilayer Perceptron (MLP) implemented with PyTorch \cite{paszke2019pytorch}. The models were trained based on the datasets sampled in the range of [0, 100]. The rest dataset within the range of [-100, 0) was used to test the generalization ability of each model. We then evaluated and compared the performance of our model and the baselines.

\begin{figure}
    \centering
    \includegraphics[scale=0.473]{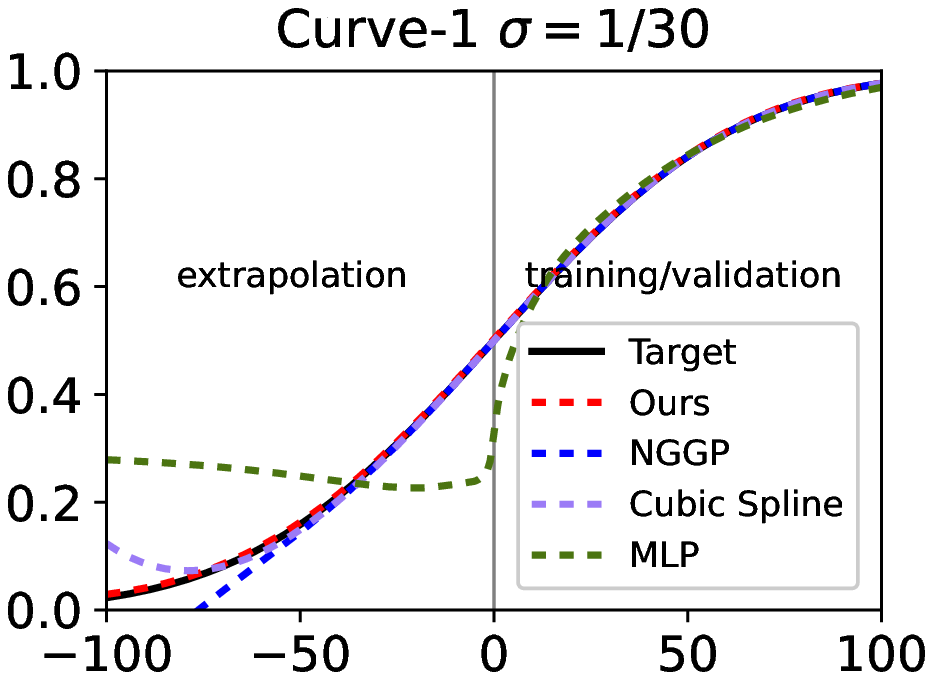}
    \hspace{-12pt}
    \includegraphics[scale=0.473]{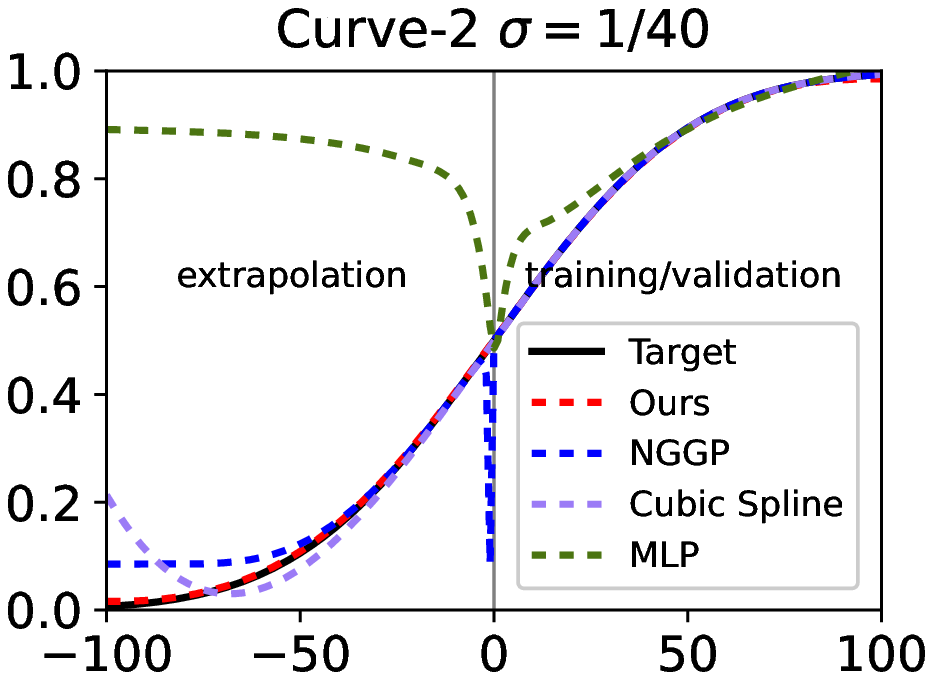}
    \hspace{-12pt}
    \includegraphics[scale=0.473]{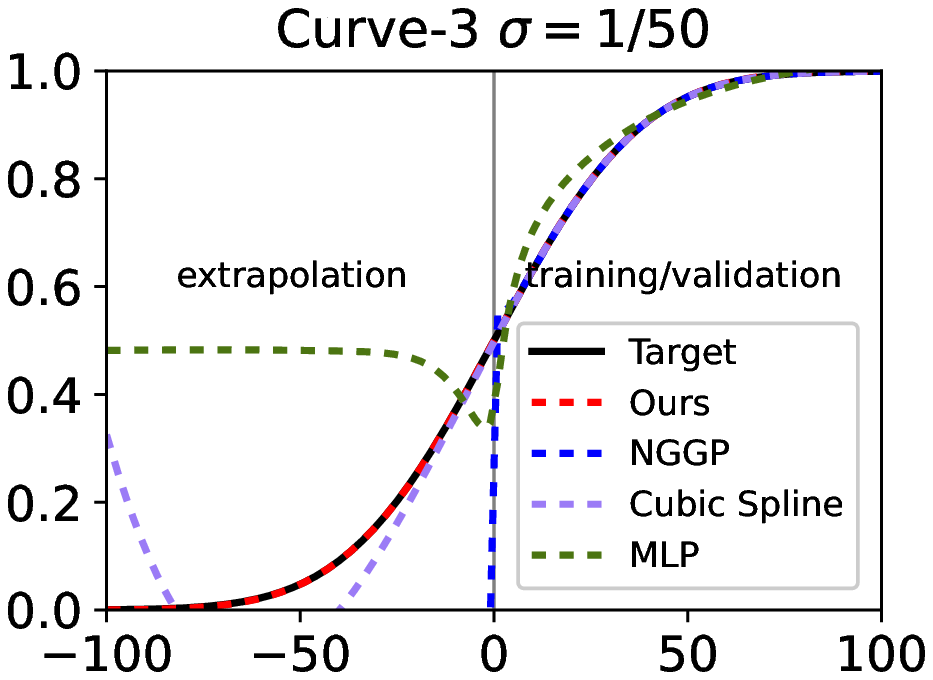}
    \caption{Generalization test experiment. The MSE metrics and the mathematical functions discovered during the generalization experiments are given in Appendix \ref{sec:gen}.}
    \label{fig:generalization}
\end{figure}

The results of the experiment, shown in Figure \ref{fig:generalization}, demonstrate that our model outperforms the baseline methods in terms of generalization ability. The curves fitted by our model exhibit better accuracy and capture the underlying patterns in the data more effectively. Moreover, the equations generated by our model possess several advantages over the baseline methods (see Appendix Table \ref{table:overall-generalization-function}), which are not only easier to calculate, but also more parsimonious in terms of their mathematical expressions.

\section{Ablation Study}

\begin{wraptable}[12]{r}{0.72\textwidth}
\vspace{-48pt}
\caption{Recovery rate (\%) of the ablation study on the LiverMore dataset. The specific recovery rate values are further shown in Appendix \ref{sec:abl}.} \label{table:ablation}
\centering
\vspace{6pt}
{\footnotesize
  \begin{tabular}{cccccc}
    \toprule
    Equation&Ours&Model A&Model B&Model C&Model D\\
    \midrule
    $\mathrm{sin}(x_1^2)\mathrm{cos}(x_1)-2$&\textbf{100}&\textbf{100}&\textbf{100}&6&\textbf{100}\\
    $\mathrm{sin}(x_1^2)\mathrm{cos}(x_1)-5$&\textbf{100}&89&60&0&\textbf{100}\\
    $\mathrm{sin}(x_1^3)\mathrm{cos}(x_1^2)-1$&\textbf{55}&20&0&0&\textbf{55}\\
    $\mathrm{sinh}(x_1)$&\textbf{100}&\textbf{100}&\textbf{100}&\textbf{100}&10\\
    $\mathrm{cosh}(x_1)$&\textbf{100}&\textbf{100}&\textbf{100}&\textbf{100}&3\\
    $\sum_{k=1}^9x^k$&\textbf{100}&83&\textbf{100}&88&\textbf{100}\\
    $x_1^{1/3}$&\textbf{100}&\textbf{100}&\textbf{100}&67&\textbf{100}\\
    $x_1^{2/5}$&\textbf{100}&\textbf{100}&\textbf{100}&12&\textbf{100}\\
    \midrule
    \textbf{Average:}&\textbf{97.95}&94.36&93.64&80.45&89.45\\
    \bottomrule
\end{tabular}}
\normalsize
\end{wraptable}

We conducted a series of ablation experiments on the LiverMore dataset, testing the ablation of double Q-learning (Model A), the ablation of the MCTS algorithm (Model B), and the ablation of the search form (Model C), and the ablation of the pre-processing step (Model D). We list some of the expressions affected by the performance of the model in Table \ref{table:ablation}. 

When the double Q-learning module is removed, Model A with only MCTS experiences a decrease in knowledge from previous iterations. This results in reduced search efficiency but increased diversity. As a result, we observe a decrease in performance for equations like $\sum_{k=1}^9x^k$, while equations like $\mathrm{sin}(x_1^3)\mathrm{cos}(x_1^2)-1$ show improved performance. On the other hand, when the MCTS module is removed, Model B with pure double A-learning tends to overfit more quickly. Consequently, it struggles to produce the most challenging equations, such as $\mathrm{sin}(x_1^3)\mathrm{cos}(x_1^2)-1$. Similarly, the absence of the expression form search module in Model C limits its ability to discover complex expressions with simple forms, such as $x_1^{1/3}$ and $\mathrm{sin}(x_1^3)\mathrm{cos}(x_1^2)-1$. Lastly, Model D, without the preprocessing module, suffers a significant reduction in its ability to search for odd and even functions like $\mathrm{sinh}(x_1)$.

These observations highlight the importance and usefulness of all the modules in our approach. Each module contributes to the overall performance and enables the model to tackle different types of equations effectively.

\section{Conclusion}
We have proposed a novel model that integrates reinforcement learning techniques, GP, and a modulated sub-tree discovery block to improve the search process for mathematical expressions. Our model outperforms state-of-the-art baselines in accurately recovering the exact representation of the baseline tasks and physical equation finding and demonstrates superior generalization capabilities. However, one limitation of our current model is the lack of flexibility in setting the expression form, which restricts its adaptability to different problem domains. We anticipate future advancements in more flexible methods, potentially incorporating neural networks to generate slots for symbolic regression. Furthermore, we believe that our approach has the potential to be extended to other domains, such as reinforcement learning control tasks. By applying our method to diverse areas, we aim to enhance the performance and applicability of symbolic regression techniques.

\section*{Acknowledgement}
The source data and codes used will be posted upon final publication of the paper. The work is supported by the National Natural Science Foundation of China (No. 92270118 and No. 62276269) and the Beijing Outstanding Young Scientist Program (No. BJJWZYJH012019100020098). Y.L. and H.S. would like to acknowledge the support from the Fundamental Research Funds for the Central Universities. 

\bibliographystyle{unsrt}
\begin{small}
\bibliography{main}
\end{small}

\appendix

\setcounter{page}{1}
\renewcommand{\thepage}{S-\arabic{page}}

\clearpage
\section*{APPENDIX}

\setcounter{figure}{0}
\setcounter{table}{0}
\renewcommand{\thefigure}{S\arabic{figure}}
\renewcommand{\thetable}{S\arabic{table}}

\section{Model Setting}
\label{sec:model}
In this section, we give more details about the settings of our models. The full set of hyperparameters can be seen in Table \ref{table:hyperparameters}.

\begin{table*}[b!]
\centering
  \caption{Hyperparameters of our model}
 \vspace{6pt}
  \begin{tabular}{ccc}
    \toprule
    Name&Abbreviation&Value\\
    \midrule
    \multicolumn{3}{c}{RL parameters}\\
    \midrule
    Minimum expression lengths&$l_{min}$&$4$\\
    Maximum expression lengths&$l_{max}$&$35$\\
    Maximum number of parameters&$c_{max}$&$10$\\
    Length discount rate &$\eta$&$0.99$\\
    Training rounds &$t_r$&$50$\\ 
    UCT constant &$c$& $\sqrt{2}$\\ 
    Minimum selected times &$n_0$& $3$\\
    Learning rate of double Q-learning &$lr$&$10^{-3}$\\
    %Stop searching reward&$\mathcal{R}_{stop}$&$10^{-10}$\\
    \midrule
    \multicolumn{3}{c}{Genetic Programming parameters}\\
    \midrule
    GP rounds &$t_{gp}$&$30$\\ 
    GP population &$p_{gp}$&$500$\\ 
    GP number of best expressions  &$l_{b}$&$20$\\
    GP Mate rate&$p_{mate}$&$0.5$\\
    GP Mutate rate&$p_{mutate}$&$0.5$\\
    \midrule
    \multicolumn{3}{c}{MSDB parameters}\\
    \midrule
    Error of Symmetry&$E_{sym}$&$10^{-5}$\\
    Selection ratio&$k_s$&$0.1$\\
    Expression percentage ratio&$k_p$&$0.1$\\
    Maximum select number&$N$&$5$\\
    \bottomrule
\end{tabular}
\label{table:hyperparameters}
\end{table*}

\textbf{Expression constraint}: 
In this study, we incorporate a prior constraint inspired by the DSR method to effectively reduce the search space for expressions. The following constraints are applied:
\begin{itemize}
    \item \textbf{Length constraint}: The length of expressions is restricted within pre-defined minimum and maximum values. If the current length falls below the minimum threshold, variables ( $x$, $y$, $C$, etc. ) and parameters will not be generated. Conversely, if the current length, combined with the number of nodes to be generated, reaches the maximum length, only these nodes will be considered.
    
    \item \textbf{Unary operator constraint}: The direct successor node of a unary operator should not be the inverse of that same operator. This constraint ensures that the generated expressions adhere to the intended structure and prevent redundant combinations.

    \item \textbf{Trigonometric function constraint}: The successor node of a trigonometric function node should not be another trigonometric function. This constraint prevents the generation of expression structures that lead to unnecessary complexity or redundancy.

    \item \textbf{Maximum parameter limit}: A specified maximum number of parameters is imposed to control the complexity of the expressions and prevent overfitting.
\end{itemize}

By applying these expression constraints, we aim to enhance the search efficiency and guide the generation of meaningful expressions that align with the desired properties of the target problem.

\section{Basic Benchmark Result}
\label{sec:basic dataset}
\subsection{Baseline Setting}
\label{sec:baseline}
In this section, we give more details about the settings of baselines( SPL, NGGP, DSR, GP ). The full set of hyperparameters can be seen below.

\begin{itemize}
    \item \textbf{SPL}: In line with the original paper, we maintain the same parameter settings for SPL. The discount rate is set to $\eta=0.9999$, and the candidate operators include addition ($+$), subtraction ($-$), multiplication ($\times$), division ($\div$), cosine ($\cos(\cdot)$), sine ($\sin(\cdot)$), exponential ($\exp(\cdot)$), natural logarithm ($\log(\cdot)$), and square root ($\sqrt{\cdot}$). Other parameter values are as follows: Maximum Module Transplantation: $20$, Episodes Between Module Transplantation: $50000$, Maximum Tree Size: $50$, and Maximum Augmented Grammars: $5$.

    \item \textbf{DSR/NGGP}: In our study, we adopt the standard parameter configurations as provided in the publicly available implementation of Deep Symbolic Optimization (DSO). This approach entails adjusting two primary hyperparameters. The entropy coefficient is set $\lambda H=0.05$ and the risk factor is set $\epsilon = 0.005$. Candidate operators are the same as those employed in the SPL. Additionally, NGGP incorporates other hyperparameters related to hybrid methods based on genetic programming. The specific values are listed in Table \ref{table:hyperparameters2}.

    \item \textbf{Genetic Programming (GP)}: We employ the gplearn library for GP-based methods. The hyperparameters for genetic programming are identical to those presented in Table \ref{table:hyperparameters2}.
    
\end{itemize}
\begin{table*}[t!]
\centering
  \caption{Genetic Programming Hyperparameters on baselines}
 \vspace{6pt}
  \begin{tabular}{cc}
    \toprule
    Name&Value\\
    \midrule
    Rounds &$20$\\ 
    Population &$1000$\\ 
    Mate rate&$0.5$\\
    Mutate rate&$0.5$\\
    %Const range&$(-5, 5)$\\
    \bottomrule
\end{tabular}
\label{table:hyperparameters2}
\end{table*}
\subsection{Nyugen Benchmark Result}
In this section, we provide additional details about the results obtained from the Nyugen and Nyugen$^c$ Benchmark experiment.

By referring to Table \ref{table:overall-Nyugen}, readers can obtain more detailed information about the performance of each model on each expression, their comparative analysis, and any other relevant insights derived from the experiment.

\begin{table*}[t!]
\centering
  \caption{Average Recovery Rate (\%) of the Nyugen Benchmark over $100$ parallel runs}
 \vspace{6pt}
  \begin{tabular}{ccccccc}
    \toprule
    Name&Equation&Ours&SPL&NGGP&DSR&GP\\
    \midrule
    Nguyen-1&$x_1^3+x_1^2+x_1$&\textbf{100}&\textbf{100}&\textbf{100}&\textbf{100}&99\\
    Nguyen-2&$x_1^4+x_1^3+x_1^2+x_1$&\textbf{100}&\textbf{100}&\textbf{100}&\textbf{100}&90\\
    Nguyen-3&$x_1^5+x_1^4+x_1^3+x_1^2+x_1$&\textbf{100}&\textbf{100}&\textbf{100}&\textbf{100}&34\\
    Nguyen-4&$x_1^6+x_1^5+x_1^4+x_1^3+x_1^2+x_1$&\textbf{100}&99&\textbf{100}&\textbf{100}&54\\
    Nguyen-5&$\mathrm{sin}(x_1^2)\mathrm{cos}(x_1)-1$&\textbf{100}&95&80&72&12\\
    Nguyen-6&$\mathrm{sin}(x_1)+\mathrm{sin}(x_1+x_1^2)$ &\textbf{100}&\textbf{100}&\textbf{100}&\textbf{100}&11\\
    Nguyen-7&$\mathrm{log}(x_1+1)+\mathrm{log}(x_1^2+1)$ &\textbf{100}&\textbf{100}&\textbf{100}&35&17\\
    Nguyen-8&$\sqrt{x_1}$ &\textbf{100}&\textbf{100}&\textbf{100}&96&76\\
    Nguyen-9 &$\mathrm{sin}(x_1)+\mathrm{sin}(x_2^2)$ &\textbf{100}&\textbf{100}&\textbf{100}&\textbf{100}&86\\
    Nguyen-10&$\mathrm{sin}(x_1)\mathrm{cos}(x_2)$  &\textbf{100}&\textbf{100}&\textbf{100}&\textbf{100}&13\\
    Nguyen-11 &$x_1^{x_2}$ &\textbf{100}&\textbf{100}&\textbf{100}&\textbf{100}&\textbf{100}\\
    Nguyen-12 &$x_1^4-x_1^3-0.5x_2^2+x_2$ &\textbf{100}&28&21&0&0\\
    \midrule
    Nguyen-$1^c$&$3.39x_1^3+2.12x_1^2+1.78x_1$&\textbf{100}&\textbf{100}&\textbf{100}&\textbf{100}&0\\
    Nguyen-$2^c$&$0.48x_1^4+3.39x_1^3+2.12x_1^2+1.78x_1$&\textbf{100}&94&\textbf{100}&\textbf{100}&0\\
    Nguyen-$5^c$&$\mathrm{sin}(x_1^2)\mathrm{cos}(x_1)-0.75$&\textbf{100}&95&98&0&1\\
    Nguyen-$7^c$&$\mathrm{log}(x_1+1.4)+\mathrm{log}(x_1^2+1.3)$ &\textbf{100}&0&\textbf{100}&93&2\\
    Nguyen-$8^c$&$\sqrt{1.23x_1}$ &\textbf{100}&\textbf{100}&\textbf{100}&\textbf{100}&56\\
    Nguyen-$9^c$ &$\mathrm{sin}(1.5x_1)+\mathrm{sin}(0.5x_2^2)$ &\textbf{100}&98&96&0&0\\
    Nguyen-$10^c$&$\mathrm{sin}(1.5x_1)\mathrm{cos}(0.5x_2)$  &\textbf{100}&0&\textbf{100}&\textbf{100}&0\\
    \midrule
    &\textbf{Average:}&\textbf{100.00}&84.68&94.47&78.74&29.00\\
    \bottomrule
\end{tabular}
\label{table:overall-Nyugen}
\end{table*}

\subsection{LiverMore Benchmark Result}
In this section, we provide additional details about the results obtained from the LiverMore Benchmark experiment.

By referring to Table \ref{table:overall-LiverMore}, readers can obtain more detailed information about the performance of each model on each expression, their comparative analysis, and any other relevant insights derived from the experiment.

\begin{table}[t!]
\centering
  \caption{Average Recovery Rate (\%) of the LiverMore Benchmark over $100$ parallel runs}
 \vspace{6pt}
  \begin{tabular}{ccccccc}
    \toprule
    Name&Equation&Ours&SPL&NGGP&DSR&GP\\
    \midrule
    Livermore-1&$1/3+x_1+\mathrm{sin}(x_1)$ &\textbf{100}&94&\textbf{100}&67&\textbf{100}\\
    Livermore-2&$\mathrm{sin}(x_1^2)\mathrm{cos}(x_1)-2$ &\textbf{100}&29&61&26&1\\
    Livermore-3&$\mathrm{sin}(x_1^3)\mathrm{cos}(x_1^2)-1$ &\textbf{55}&50&2&0&0\\
    Livermore-4&$\mathrm{log}(x_1+1)+\mathrm{log}(x_1^2+x_1)+\mathrm{log}(x_1)$ &\textbf{100}&61&\textbf{100}&72&\textbf{100}\\
    Livermore-5&$x_1^4-x_1^3+x_1^2-x_2$ &\textbf{100}&\textbf{100}&\textbf{100}&55&\textbf{100}\\
    Livermore-6&$4x_1^4+3x_1^3+2x_1^2+x_1$ &\textbf{100}&8&\textbf{100}&\textbf{100}&\textbf{100}\\
    Livermore-7&$\mathrm{sinh}(x_1)$ &\textbf{100}&18&24&0&0\\
    Livermore-8&$\mathrm{cosh}(x_1)$ &\textbf{100}&6&30&0&0\\
    Livermore-9&$\sum_{i=1}^9x_1^i$ &\textbf{100}&21&99&18&0\\
    Livermore-10&$6\mathrm{sin}(x_1)\mathrm{cos}(x_2)$ &\textbf{100}&75&\textbf{100}&70&23\\
    Livermore-11&$(x_1^2x_2^2)/(x_1+x_2)$ &\textbf{100}&0&\textbf{100}&78&95\\
    Livermore-12&$x_1^5/x_2^3$ &\textbf{100}&\textbf{100}&\textbf{100}&13&\textbf{100}\\
    Livermore-13&$x_1^{1/3}$ &\textbf{100}&12&\textbf{100}&59&0\\
    Livermore-14&$x_1^3+x_1^2+x_1+\mathrm{sin}(x_1)+\mathrm{sin}(x_1^2)$ &\textbf{100}&\textbf{100}&\textbf{100}&91&\textbf{100}\\
    Livermore-15&$x_1^{1/5}$ &\textbf{100}&0&\textbf{100}&28&2\\
    Livermore-16&$x_1^{2/5}$ &\textbf{100}&0&26&0&0\\
    Livermore-17&$4\mathrm{sin}(x_1)\mathrm{cos}(x_2)$ &\textbf{100}&89&\textbf{100}&\textbf{100}&84\\
    Livermore-18&$\mathrm{sin}(x_1^2)\mathrm{cos}(x_1)-5$ &\textbf{100}&18&33&37&0\\
    Livermore-19&$x_1^5+x_1^4+x_1^2+x_1$ &\textbf{100}&89&\textbf{100}&\textbf{100}&\textbf{100}\\
    Livermore-20&$\mathrm{exp}(-x_1^2)$ &\textbf{100}&\textbf{100}&\textbf{100}&\textbf{100}&\textbf{100}\\
    Livermore-21&$\sum_{i=1}^8x_1^i$ &\textbf{100}&52&\textbf{100}&13&12\\
    Livermore-22&$\mathrm{exp}(-0.5x_1^2)$ &\textbf{100}&\textbf{100}&\textbf{100}&82&\textbf{100}\\
    \midrule
    &\textbf{Average}&\textbf{97.95}&51.0&80.68&50.41&50.77\\
    \bottomrule
\end{tabular}
\label{table:overall-LiverMore}
\end{table}

\subsection{R Benchmark Result}
In this section, we provide additional details about the results obtained from the R Rational Benchmark experiment.

By referring to Table \ref{table:overall-R}, readers can obtain more detailed information about the performance of each model on each expression, their comparative analysis, and any other relevant insights derived from the experiment.

\begin{table*}[t!]
\centering
  \caption{Average Recovery Rate (\%) of the R Rational Benchmark over $100$ parallel runs}
 \vspace{6pt}
  \begin{tabular}{ccccccc}
    \toprule
    Name&Equation&Ours&SPL&NGGP&DSR&GP\\
    \midrule
    R$^0$-1&$(x_1+1)^3/(x_1^2-x_1+1)$&5&0&\textbf{15}&0&0\\
    R$^0$-2&$(x_1^5-3x_1^3+1)/(x_1^2+1)$&\textbf{80}&0&40&0&0\\
    R$^0$-3&$(x_1^5+x_1^6)/(x_1^4+x_1^3+x_1^2+x_1+1)$&\textbf{100}&0&\textbf{100}&0&0\\
    R$^*$-1&$(x_1+1)^3/(x_1^2-x_1+1)$&\textbf{48}&0&2&0&0\\
    R$^*$-2&$(x_1^5-3x_1^3+1)/(x_1^2+1)$&\textbf{89}&0&0&0&0\\
    R$^*$-3&$(x_1^5+x_1^6)/(x_1^4+x_1^3+x_1^2+x_1+1)$&\textbf{91}&0&3&0&0\\
    \midrule
    &\textbf{Average}&\textbf{68.83}&0.0&26.67&0.0&0.0\\
    \bottomrule
\end{tabular}
\label{table:overall-R}
\end{table*}

\section{Free-falling Balls Dataset Result}
\label{sec:phy}
In this section, we provide additional details about the results obtained from the Falling-Balls dataset experiment.To improve the performance of the SPL model, we incorporated the operators $\log(\cosh(\cdot))$. This addition aimed to enhance the model's ability to capture the underlying patterns in the data.

The complete results of the experiment, including the functions found, can be found in Table \ref{table:overall-physical}.

\begin{table}[t!]
\small
\centering
  \caption{Functions generated in Falling-Balls Experiment}
 \vspace{6pt}
  \begin{tabular}{ccccccc}
    \toprule
    Name&Model&Equation\\
    \midrule
\multirow{3}{*}{baseball}&Ours&$-4.43t^2+0.36\sin(t^2+1.51)^2+47.35$\\
&Model A&$0.09t^3-5.47t^2+2.47t+46.52+\cos(t^2-2.5t)^{0.5}$\\
&SPL&$-4.54t^2+ 0.625t+47.8$\\
\midrule
\multirow{3}{*}{blue basket ball}&Ours&$-1.66t^3-4.95t^2\cos(\sqrt{t})+46.46$\\
&Model A&$-0.1t^3-4.49t^2+37.54t+46.49-t(\cos(t)+36.77)$\\
&SPL&$- 0.25t^4 + t^3- 5.11t^2+46.47  $\\
\midrule
\multirow{3}{*}{bowling ball}&Ours&$-4.63t^2+\sin(0.83t)\sin(t)+46.13$\\
&Model A&$0.18t^3-6.0t^2+2.15t+45.43+|t-0.62|$\\
&SPL&$- 0.285t^3-3.82t^2  + 4.14\times 10^{-5}\exp(20.74t^2-12.45t^3) +46.1$\\
\midrule
\multirow{3}{*}{golf ball}&Ours&$-0.09t^3-4.44t^2+5.26\times10^{-5}t/\log(t)+49.51$\\
&Model A&$-2.18t^3+11.75t^2+1.96t+25.86-2.36\exp(t)+25.98\cos(t)$\\
&SPL& $ - 4.9633t^2 + \log(\cosh(t))+49.5087 $\\
\midrule
\multirow{3}{*}{green basket ball}&Ours&$46.34-4.15t^2$\\
&Model A&$-0.09t^3-4.59t^2+1.6t+45.26+(\frac{0.02\sqrt{t}}{t-\exp(\cos(t))}-t+1)\cos(t)$\\
&SPL&$-4.1465t^2 + 45.9087 +\log(\cosh(1))$\\
\midrule
\multirow{3}{*}{tennis ball}&Ours&$47.78\cos(0.43t-0.02)$\\
&Model A&$0.33t^3-4.9t^2+0.66t+47.74$\\
&SPL&$-4.0574t^2 + \log(\cosh(0.121t^3))+47.8577$\\
\midrule
\multirow{3}{*}{volleyball}&Ours&$48.15-3.67(t+0.03)^2$\\
&Model A&$1.59t^3-11.1t^2+0.93t+58.53-10.53\cos(t)$\\
&SPL&$- 3.78t^2+48.0744$\\
\midrule
\multirow{3}{*}{whiffle ball1}&Ours&$-t^2(3.83-0.31t)+47.07$\\
&Model A&$-0.08t^3-2.17t^2-1.69t+46.29+\sqrt{t+\sin(3t)}$\\
&SPL&$-t^3+4.16t^2  + 47.01\exp(-0.15t^2)$\\
\midrule
\multirow{3}{*}{whiffle ball2}&Ours&$-2.18t^2+0.1t\cos(t)+3.35\cos(t)+43.88$\\
&Model A&$0.46t^3-4.39t^2+0.19t+47.26-0.05\cos(\exp(t))$\\
&SPL&$65.86 \exp(-0.0577t^2)-18.61 $\\
\midrule
\multirow{3}{*}{yellow whiffle ball}&Ours&$(\cos(1.75t)+47.59)\cos(0.36t)$\\
&Model A&$-0.27t^3-2.58t^2-2.5t+48.25+(t+0.41)\exp(\sqrt{t+t^2-2\sqrt{t^3}})$\\
&SPL&$(148.99- 14.58t^2+48.96\log(\cosh(x)))/(\log(\cosh(t)) + 3.065)$\\
\midrule
\multirow{3}{*}{orange whiffle ball}&Ours&$-17.82t-33.11/\exp(t)^{0.5}+80.94$\\
&Model A&$0.42t^3-3.81t^2-1.4t+47.84$\\
&SPL&$-1.66t + 47.86\exp(-0.0682t^2)$\\
    \bottomrule
\end{tabular}
\label{table:overall-physical}
\end{table}

\section{Generalization Experiment Result}
\label{sec:gen}
In this section, we provide further details regarding the Generalization Benchmark on Gaussian results. 

%In the deep learning approach, we employ a Multilayer Perceptron (MLP) architecture with one input, one output, and a hidden layer ranging in size from 1 to 40. We experiment with different configurations of the hidden layer and select the model that yields the best performance. The MLP is trained using the training set, and the test set is used to evaluate the performance of each model. By varying the size of the hidden layer, we aim to find the optimal architecture that achieves the highest accuracy or lowest error on the given task. 

Each dataset is divided into three subsets: a training set, a test set, and a validation set. The training set comprises points ranging from $30$ to $80$, while the test set consists of points ranging from $10$ to $25$. The validation set covers a broader range, spanning from $0$ to $100$.

To evaluate the performance of our approach and compare it with baselines, we define the following settings for each method:
\begin{itemize}
    \item \textbf{Ours}: The training set is utilized for generating expressions and calculating the corresponding rewards. The test set is employed to evaluate the quality of the generated expressions. Finally, the validation set is employed to select the most promising expressions from the outputs.
    
    \item \textbf{NGGP}: Both the training set and the test set are used for generating expressions and computing rewards. The HallOfFame, which contains the best expressions, is then leveraged to choose expressions using the validation set.
    
    \item \textbf{Linear regression}: The training set and the test set are employed for training the linear regression model.
    
    \item \textbf{Cubic spline}: The training set and the test set are used to train the cubic spline model.
    
    \item \textbf{Deep learning}: In the deep learning approach, we employ a Multilayer Perceptron (MLP) architecture with one input, one output, and a hidden layer ranging in size from 30 to 50. We set learning rate to $10^{-3}$ and train 100 epoches. We experiment with different configurations of the hidden layer and select the model that yields the best performance. The MLP is trained using the training set, and the test set is used to evaluate the performance of each model. By varying the size of the hidden layer, we aim to find the optimal architecture that achieves the highest accuracy or lowest error on the given task. 
\end{itemize}

The full results of the generalization experiment can be found in Table \ref{table:overall-generalization}. This table presents a detailed overview of the performance of the model and the baselines. Additionally, Table \ref{table:overall-generalization-function} presents the equations discovered by the model and the baselines.

\begin{table}[t!]
\small
\centering
  \caption{Mean Squared Error (MSE) of each method and each part of the curve in the Generalization Experiment}
 \vspace{6pt}
  \begin{tabular}{cccccc}
    \toprule
    Name&Ours&NGGP&Linear&Cubic Spline&MLP\\
    \midrule
total error on curve 1 & $\mathbf{1.05\times 10^{-5}}$ &$ 0.00215$ &$ 0.0114$ &$ 0.000381$ &$ 0.0142$\\
total error on curve 2 & $\mathbf{9.79\times 10^{-6}}$ &$ 0.00163$ &$ 0.0297$ &$ 0.00162$ &$ 0.261$\\
total error on curve 3 & $\mathbf{2.61\times 10^{-7}}$ &$ 0.327$ &$ 0.0821$ &$ 0.00563$ &$ 0.0762$\\
\midrule
extrapolation error on curve 1 & $\mathbf{1.65\times 10^{-5}}$ &$ 0.00429$ &$ 0.0215$ &$ 0.000758$ &$ 0.0278$\\
extrapolation error on curve 2 & $\mathbf{1.41\times 10^{-5}}$ &$ 0.00324$ &$ 0.0565$ &$ 0.00323$ &$ 0.518$\\
extrapolation error on curve 3 & $\mathbf{2.67\times 10^{-7}}$ &$ 0.65$ &$ 0.158$ &$ 0.0112$ &$ 0.151$\\
\midrule
validation error on curve 1 &$ 4.46\times 10^{-6}$ & $\mathbf{2.59\times 10^{-10}}$ &$ 0.00129$ &$ 2.76\times 10^{-10}$ &$ 0.000532$\\
validation error on curve 2 &$ 5.45\times 10^{-6}$ & $\mathbf{1.08\times 10^{-10}}$ &$ 0.00265$ &$ 2.87\times 10^{-9}$ &$ 0.00177$\\
validation error on curve 3 &$ 2.55\times 10^{-7}$ &$ 2.94\times 10^{-5}$ &$ 0.00518$ & $\mathbf{3.75\times 10^{-8}}$ &$ 0.00106$\\
\midrule
test error on curve 1 &$ 6.95\times 10^{-6}$ &$ 9.09\times 10^{-12}$ &$ 0.000545$ & $\mathbf{<1\times 10^{-12}}$ &$ 0.000253$\\
test error on curve 2 &$ 2.25\times 10^{-7}$ & $\mathbf{<1\times 10^{-12}}$ &$ 0.00127$ & $\mathbf{<1\times 10^{-12}}$ &$ 0.00344$\\
test error on curve 3 &$ 8.8\times 10^{-8}$ & $\mathbf{<1\times 10^{-12}}$ &$ 0.00272$ & $\mathbf{<1\times 10^{-12}}$ &$ 0.00358$\\
\midrule
training error on curve 1 &$ 3.88\times 10^{-6}$ &$ 3.57\times 10^{-12}$ &$ 0.000254$ & $\mathbf{<1\times 10^{-12}}$ &$ 6.93\times 10^{-5}$\\
training error on curve 2 &$ 1.26\times 10^{-6}$ & $\mathbf{<1\times 10^{-12}}$ &$ 0.000524$ & $\mathbf{<1\times 10^{-12}}$ &$ 8.6\times 10^{-5}$\\
training error on curve 3 &$ 3.34\times 10^{-7}$ &$ 7.14\times 10^{-12}$ &$ 0.000905$ & $\mathbf{<1\times 10^{-12}}$ &$ 7.9\times 10^{-5}$\\
\bottomrule
\end{tabular}
\label{table:overall-generalization}
\end{table}

\begin{table}[t!]
\centering
  \caption{Functions generated in Generalization Experiment. Our functions are easier to calculate and shorter than NGGP's.}
 \vspace{6pt}
  \resizebox{\textwidth}{!}{
  \begin{tabular}{ccccccc}
    \toprule
    Name&Model&Equation\\
    \midrule
\multirow{3}{*}{curve-1}&Ours&$0.503+(117.088x)/(x^2+14702)$\\
&NGGP&$\cos(\exp((0.49x\log(0.028x + 29.5/(0.039x + 9.82)) - 2.78)/(0.115x - 60.5)))$\\
%&Linear&$0.58528+0.004831266x$\\
\midrule
\multirow{3}{*}{curve-2}&Ours&$(6.08x+0.785)/(0.0639x^2+615.179)+0.50003$\\
&NGGP&$\cos(2.95\exp(-0.68\exp(0.41\exp(5.5x\exp(24.67/(115.6\exp((4.75x + 13.3)/x) + 3.2))/(2x + 214.3)))))$\\
%&Linear&$0.63418 +0.00473183x$\\
\midrule
\multirow{3}{*}{curve-3}&Ours&$(x\sin(371.57/(13928/x+x))+x)/(0.0024+2x)$\\
&NGGP&$\cos(\log(1 + 1.67\exp(-1.56/(\exp(17.7\exp(\exp((-12.9 + 4.95\log(x)/x)/x))/x) - 1.16 + 0.656/x))))$\\
%&Linear&$0.71790+0.004012465x$\\
    \bottomrule
\end{tabular}
}
\label{table:overall-generalization-function}
\end{table}

\section{Ablation Experiment Result}
\label{sec:abl}
In this section, we provide more detailed information about the results obtained from the ablation experiment on LiverMore benchmark. The full results of the ablation experiment can be found in Table \ref{table:ablation-overall}. This table presents a comprehensive overview of the performance of the model under different ablation settings.
\begin{table}[t!]
\small
\centering
  \caption{Average Recovery Rate (\%) of the Ablation Experiment over $100$ parallel runs}
 \vspace{6pt}
  \begin{tabular}{ccccccc}
    \toprule
    Name&Equation&Ours&ModelA&ModelB&ModelC&ModelD\\
    \midrule
    Livermore-1&$1/3+x_1+\mathrm{\sin}(x_1)$ &\textbf{100}&\textbf{100}&\textbf{100}&\textbf{100}&\textbf{100}\\
Livermore-2&$\mathrm{sin}(x_1^2)\mathrm{cos}(x_1)-2$ &\textbf{100}&\textbf{100}&\textbf{100}&6&\textbf{100}\\
Livermore-3&$\mathrm{sin}(x_1^3)\mathrm{cos}(x_1^2)-1$ &\textbf{55}&20&0&0&\textbf{55}\\
Livermore-4&$\mathrm{log}(x_1+1)+\mathrm{log}(x_1^2+x_1)+\mathrm{log}(x_1)$ &\textbf{100}&\textbf{100}&\textbf{100}&\textbf{100}&\textbf{100}\\
Livermore-5&$x_1^4-x_1^3+x_1^2-x_2$ &\textbf{100}&\textbf{100}&\textbf{100}&\textbf{100}&\textbf{100}\\
Livermore-6&$4x_1^4+3x_1^3+2x_1^2+x_1$ &\textbf{100}&\textbf{100}&\textbf{100}&\textbf{100}&\textbf{100}\\
Livermore-7&$\mathrm{sinh}(x_1)$ &\textbf{100}&\textbf{100}&\textbf{100}&\textbf{100}&10\\
Livermore-8&$\mathrm{cosh}(x_1)$ &\textbf{100}&\textbf{100}&\textbf{100}&\textbf{100}&3\\
Livermore-9&$\sum_{i=1}^9x_1^i$ &\textbf{100}&83&\textbf{100}&88&\textbf{100}\\
Livermore-10&$6\mathrm{sin}(x_1)\mathrm{cos}(x_2)$ &\textbf{100}&\textbf{100}&\textbf{100}&\textbf{100}&\textbf{100}\\
Livermore-11&$(x_1^2x_2^2)/(x_1+x_2)$ &\textbf{100}&91&\textbf{100}&\textbf{100}&\textbf{100}\\
Livermore-12&$x_1^5/x_2^3$ &\textbf{100}&\textbf{100}&\textbf{100}&\textbf{100}&\textbf{100}\\
Livermore-13&$x_1^{1/3}$ &\textbf{100}&\textbf{100}&\textbf{100}&67&\textbf{100}\\
Livermore-14&$x_1^3+x_1^2+x_1+\mathrm{sin}(x_1)+\mathrm{sin}(x_1^2)$ &\textbf{100}&\textbf{100}&\textbf{100}&\textbf{100}&\textbf{100}\\
Livermore-15&$x_1^{1/5}$ &\textbf{100}&\textbf{100}&\textbf{100}&97&\textbf{100}\\
Livermore-16&$x_1^{2/5}$ &\textbf{100}&\textbf{100}&\textbf{100}&12&\textbf{100}\\
Livermore-17&$4\mathrm{sin}(x_1)\mathrm{cos}(x_2)$ &\textbf{100}&\textbf{100}&\textbf{100}&\textbf{100}&\textbf{100}\\
Livermore-18&$\mathrm{sin}(x_1^2)\mathrm{cos}(x_1)-5$ &\textbf{100}&89&90&0&\textbf{100}\\
Livermore-19&$x_1^5+x_1^4+x_1^2+x_1$ &\textbf{100}&\textbf{100}&\textbf{100}&\textbf{100}&\textbf{100}\\
Livermore-20&$\mathrm{exp}(-x_1^2)$ &\textbf{100}&\textbf{100}&\textbf{100}&\textbf{100}&\textbf{100}\\
Livermore-21&$\sum_{i=1}^8x_1^i$ &\textbf{100}&\textbf{100}&\textbf{100}&\textbf{100}&\textbf{100}\\
Livermore-22&$\mathrm{exp}(-0.5x_1^2)$ &\textbf{100}&\textbf{100}&\textbf{100}&\textbf{100}&\textbf{100}\\
\midrule
    &\textbf{Average}&\textbf{97.95}&94.68&95.0&80.45&89.45\\
    \bottomrule
\end{tabular}
\label{table:ablation-overall}
\end{table}

\end{document}